\documentclass{article}
\usepackage{ijcai25}
\usepackage{times}
\usepackage{soul}
\usepackage{url}
\usepackage[hidelinks]{hyperref}
\usepackage[utf8]{inputenc}
\usepackage[small]{caption}
\usepackage{graphicx}
\usepackage{amsmath}
\usepackage{amsthm}
\usepackage{booktabs}
\usepackage{algorithm}
\usepackage{algorithmic}
\usepackage[switch]{lineno}
\usepackage{amssymb}
\usepackage{multirow}
\usepackage{graphicx}
\usepackage{bm}
\usepackage{pifont}

\usepackage{tikz}
\usetikzlibrary{shapes.geometric, arrows.meta, positioning, shadows}
\usepackage{tikz-qtree}
\usetikzlibrary{decorations}
\usetikzlibrary{mindmap, shadows, trees}

\title{Survey of Adversarial Robustness in Multimodal Large Language Models}

\author{
    Author Name
    \affiliations
    Affiliation
    \emails
    email@example.com
}

\author{
Chengze Jiang$^1$\and
Zhuangzhuang Wang$^1$\and
Minjing Dong$^3$\and
Jie Gui$^{*1,2}$\and
\affiliations
$^1$School of Cyber Science and Engineering, Southeast University, Nanjing, China\\
$^2$Purple Mountain Laboratories, Nanjing, China\\
$^3$Department of Computer Science, City University of Hong Kong, Hong Kong, China\\
\emails
czjiang@seu.edu.cn,
220235252@seu.edu.cn,
minjdong@cityu.edu.hk,
guijie@seu.edu.cn
}

\begin{document}

\maketitle

\begin{abstract}
Multimodal Large Language Models (MLLMs) have demonstrated exceptional performance in artificial intelligence by facilitating integrated understanding across diverse modalities, including text, images, video, audio, and speech. However, their deployment in real-world applications raises significant concerns about adversarial vulnerabilities that could compromise their safety and reliability. Unlike unimodal models, MLLMs face unique challenges due to the interdependencies among modalities, making them susceptible to modality-specific threats and cross-modal adversarial manipulations. This paper reviews the adversarial robustness of MLLMs, covering different modalities. We begin with an overview of MLLMs and a taxonomy of adversarial attacks tailored to each modality. Next, we review key datasets and evaluation metrics used to assess the robustness of MLLMs. After that, we provide an in-depth review of attacks targeting MLLMs across different modalities. Our survey also identifies critical challenges and suggests promising future research directions. 
\end{abstract}

\section{Introduction}
Multimodal large language models (MLLMs) have driven transformative advancements in artificial intelligence by enabling the seamless integration of diverse data modalities \cite{zhang2024vision}. Unlike unimodal models that are limited to processing a single input modality, MLLMs empower a broad range of applications, from image captioning and visual question answering to video summarization and audio-visual speech recognition \cite{sun2024trustnavgpt}. Grounded in their ability to understand and integrate multimodal information, MLLMs enable the comprehension and generation of content across diverse data types, offering a holistic understanding that captures the intricate relationships between modalities and reflects real-world interactions \cite{IJCAISur} \cite{zhao2024evaluating}.
\par
Despite these advances, existing research reveals that MLLMs remain vulnerable to adversarial attacks \cite{liu2024survey}. While multimodal integration is a key strength, it simultaneously expands the attack surface, introducing new vectors for manipulation in both modality-specific and cross-modal scenarios \cite{qi2024visual}. Although various studies have explored adversarial strategies within individual modalities, no unified perspective has emerged to address the complexities of MLLMs comprehensively \cite{liu2024survey}. Besides, adversarial threats can arise not only during inference but also in the training phase \cite{xu2024shadow}. These vulnerabilities introduce significant risks by embedding latent threats into MLLMs, which are posed by adversarial attacks and are further amplified by the growing deployment of MLLMs in high-stakes applications, including autonomous systems \cite{fu2024pg}, medical diagnostics \cite{fan2024unbridled}, and content moderation \cite{gong2023figstep}. Given these multifaceted threats, the question of developing robust defense mechanisms becomes paramount. However, current defenses often lag behind the sophistication of emerging attacks, as they primarily target unimodal adversarial scenarios and fail to account for the complex dynamics of multimodal data integration. 
\par
Considering the security concerns arising from the robustness risks of MLLMs, as well as the rapid advancements in this field, there is an urgent requirement for a systematic investigation that not only informs but also guides future research. To address this necessity, we present a comprehensive survey that examines adversarial risks and defenses of MLLMs across a variety of modalities. Unlike prior studies that predominantly focus on image-text models, our survey expands the scope to encompass additional modalities and their corresponding defense strategies, thus offering a broader perspective on the adversarial robustness of MLLMs. Specifically, we categorize these approaches into four distinct modalities, covering image, video, audio, and speech. This categorization provides a comprehensive overview and an in-depth analysis of the threats and corresponding countermeasures in each domain.
\\
\textbf{Goals of our survey.}
We aim to (i) summarize representative datasets and metrics used to evaluate MLLMs adversarial robustness across different modalities; (ii) develop a taxonomy of adversarial robustness in MLLMs from the perspective of input modalities, introducing attack strategies and robustness evaluation methods tailored to each modality; and (iii) identify open challenges to inspire future research and enhance the adversarial robustness of MLLMs.
\\
\textbf{Difference between our and previous surveys.} 
As presented in Table. \ref{RiccatiCompare}, \cite{chowdhury2024breaking} focus on text-based attacks in LLMs. \cite{IJCAISur} provide a macroscopic review of LVLM safety issues but limit their scope to vision and text modalities with preliminary taxonomy. While \cite{liu2024survey} subsequently delivers a detailed analysis of vision-text attacks, their work neglects other modalities. Similarly, \cite{fan2024unbridled} narrow their analyses to image-based attacks. These reviews lack a systematic review of the adversarial robustness of MLLMs across diverse modalities, which is a crucial aspect of MLLMs. Therefore, we present a comprehensive survey and taxonomy of the adversarial robustness of MLLMs, encompassing various modalities. Our proposed taxonomy, illustrated in Fig. \ref{OverallTaxonomy}, provides a structured overview of adversarial robustness in MLLMs from a modal perspective.

\begin{table}[t]\centering
    \setlength{\tabcolsep}{1.4mm}{
    \begin{tabular}[l]{@{}l| c c c}
    \toprule[2pt]
    Paper &Attack &Defense &Modal \\
    \toprule[0.5pt]
        \cite{chowdhury2024breaking} & $\checkmark$ & & Text\\
        \cite{fan2024unbridled}      & $\checkmark$ & & Image, Text\\
        \cite{IJCAISur}              & $\checkmark$ & $\checkmark$ &Image, Text\\
        \cite{liu2024survey}         & $\checkmark$ & &Image, Text\\
        Ours                         & $\checkmark$ &$\checkmark$ & Multimodal\\
    \toprule[2pt]
\end{tabular}
\caption{Comparisons Existing Survey for Adversarial Robustness of MLLMs}\label{RiccatiCompare}
}
\end{table}

\section{Preliminaries and Background}
\subsection{Overview of MLLMs}
MLLMs integrate multiple modalities into a unified framework, enabling cross-modal reasoning for tasks like captioning, retrieval, and translation \cite{carlini2024aligned}. These models leverage large language model backbones, such as GPT-based models, to process textual data, while specialized encoders extract features from other modalities\cite{qifine}. MLLMs enhance their ability to understand and generate information across diverse data sources by capturing intricate interdependencies among modalities \cite{zhu2024llava}. To achieve integration, MLLMs rely on fusion modules that align and combine multimodal features, often mapping them into shared latent spaces for unified representation. This facilitates interaction and alignment between modalities, enabling cohesive understanding and generation. The ability to unify and process diverse modalities not only broadens the scope of achievable tasks but also establishes MLLMs as a crucial technology for advancing cross-modal AI applications \cite{rafailov2024direct}.

\subsection{Introduction of Different Attack Methods}
\subsubsection{Adversarial Attack}
Adversarial attacks on MLLMs aim to exploit their vulnerabilities by introducing imperceptible perturbations to input data, causing the model to produce incorrect or harmful outputs. These attacks pose significant threats to applications, and the primary goal of attacks is to manipulate the model’s responses while remaining undetectable to human observers.
\par
Given an MLLM $M$ that takes an image $x$ and a textual prompt $t_\text{in}$ and model $M$ produces an output $t_\text{out}=M(x,t_\text{in})$. The objective of adversarial attack is to generate perturbed inputs $x^\text{adv}$ and $t^\text{adv}$, satisfies that:
\begin{subequations}
    \begin{align}
        & ~~~~~~~~ M(x^\text{adv}, t^\text{adv}_{\text{in}})\ne M(x, t_\text{in}), \label{MTCEM1} \\
        &~\text{s.t.} ~~~ \|x^\text{adv}-x\|_p\leq \epsilon, \|t^\text{adv}-t_\text{in}\|_p\leq\epsilon, \label{MTCEM2}
    \end{align}
\end{subequations}
where $\epsilon$ is the perturbation budget, ensuring perceptibility.
\par
In adversarial image attacks, subtle perturbations are introduced to visual inputs, leading to misclassification or incorrect textual descriptions \cite{zhao2024evaluating}. Attacks targeting visual inputs include gradient-based methods, which iteratively perturb pixels to maximize classification errors \cite{schlarmann2023adversarial}. In contrast, diffusion-based attacks further refine visual perturbations, enhancing attack effectiveness across diverse prompts and models \cite{guo2024efficient}. Text-based adversarial attacks manipulate textual inputs to mislead the model into producing incorrect outputs while maintaining syntactic plausibility \cite{dong2023robust} \cite{wang2024stop}. Furthermore, cross-modal attacks extend this approach by jointly perturbing both visual and textual inputs. The mathematical formulation for these attacks can be expressed as follows:
\begin{subequations}
    \begin{align}
    &\max_{\delta_x, \delta_t} L(M(x + \delta_x, t_{\text{in}} + \delta_t), t_{\text{target}}), \\
    &\text{s.t. } \|\delta_x\|_p \leq \epsilon_x, \|\delta_t\|_p \leq \epsilon_t
    \end{align}
\end{subequations}
where $\delta_x$ and $\delta_t$ represent the adversarial perturbations applied to the visual and textual inputs, respectively, while $\epsilon_x$ and $\delta_t$ define the perturbation imperceptibility.
\subsubsection{Jailbreak Attacks}
Jailbreak attacks aim to circumvent safety mechanisms by exploiting vulnerabilities in both textual and visual inputs, leading to the generation of harmful content. Attackers use Image Jailbreaking Prompts and Perturbed Jailbreaking Prompts to craft adversarial examples that bypass security constraints across multiple prompts and models \cite{niu2024jailbreaking} \cite{wang2024white}. Besides, Infectious Jailbreak attacks spread malicious inputs rapidly through agent interactions, causing large-scale security breaches \cite{guagent}.

\subsubsection{Data Integrity Attack}
Data integrity attacks aim to compromise reliability and trustworthiness by injecting malicious or manipulated data, ultimately leading to incorrect, biased, or even harmful outputs \cite{xu2024shadow}. These attacks can occur during both training and inference phases, with methods such as data poisoning, where adversaries introduce carefully crafted corrupted examples into the training set to influence the model’s behavior and degrade its performance over time \cite{liang2024vltrojan}. Another prevalent method is backdoor attacks, which embed hidden triggers within the data that remain dormant under normal conditions but activate malicious behaviors when specific inputs or prompts are provided \cite{lu2024test} \cite{ni2024physical}. 

\subsection{Dataset}
Various datasets across different modalities are established to evaluate the performance of MLLMs. For image-based MLLMs, commonly used datasets include MS-COCO for image captioning \cite{lin2014microsoft}, Visual7W for visual question answering, and ScienceQA for multimodal science reasoning \cite{lu2022learn}. Additionally, benchmarks such as SafetyBench and MM-SafetyBench assess model robustness against adversarial inputs and safety risks \cite{zhang2023safetybench}. In the video domain, datasets like ActivityNet-200 \cite{activitynet} and MSVD-QA \cite{xu2017video} focus on evaluating models' ability to capture temporal coherence and event alignment, which are critical for understanding dynamic visual content. For audio-based MLLMs, LibriSpeech \cite{panayotov2015librispeech} is the benchmark for speech recognition and speaker verification tasks, while LlamaPartialSpoof \cite{luong2024llamapartialspoof} is used to evaluate robustness against spoofing attacks in speech-based applications. The details of datasets are shown in Table \ref{DatasetSummary}.

\begin{table}[t]\centering
\setlength{\tabcolsep}{1.0mm}{
\begin{tabular}{l|c|c|c|c|c}
\toprule[2pt]
\textbf{Dataset} & \textbf{I} & \textbf{T} & \textbf{V} & \textbf{A} & \textbf{S} \\
\toprule[1pt]
MS-COCO \cite{lin2014microsoft} & $\checkmark$ &  &  & \\
Flickr30K \cite{flickr30k} & $\checkmark$ & $\checkmark$ &  & \\
VizWiz \cite{VizWiz} & $\checkmark$ & $\checkmark$ &  & \\
VQA-v2 \cite{VQA-v2} & $\checkmark$ & $\checkmark$ &  & \\
A-OKVQA \cite{ok-vqa} & $\checkmark$ & $\checkmark$ &  & \\
ScienceQA \cite{lu2022learn} & $\checkmark$ & $\checkmark$ &  & \\
\hline
ActivityNet \cite{activitynet} &  & $\checkmark$ & $\checkmark$ & \\
MSVD-QA \cite{xu2017video} &  & $\checkmark$ & $\checkmark$ & \\
VideoQA \cite{xiao2024videoqa} &  & $\checkmark$ & $\checkmark$ & \\
EditVid-QA \cite{xu2024beyond} &  & $\checkmark$ & $\checkmark$ & \\
AV-Deepfake1M \cite{cai2024av} &  &  & $\checkmark$ & $\checkmark$ &\\
LlamaPartialSpoof \cite{luong2024llamapartialspoof} &  &  & & &$\checkmark$ \\
\toprule[2pt]
\end{tabular}
}
\caption{Dataset Summary. I, T, V, A, and S represent image, text, video, audio, and speech modalities, respectively.}\label{DatasetSummary}
\end{table}


\subsection{Adversarial Risk of MLM} 
Despite significant advancements in both open- and closed-source MLLMs, ensuring resilience to adversarial perturbations and maintaining safety across modalities remains an open research challenge \cite{zhao2024evaluating}. This challenge is particularly pronounced in multimodal information, where attacks can target a single modality (e.g., noise in images or manipulations in text) or disrupt cross-modal alignments \cite{liu2025mm}, \cite{qi2024visual}. Moreover, unlike conventional adversarial attacks that typically rely on gradient-based methods to generate adversarial examples, attacks on MLLMs are more diverse, and there is currently no consensus on the definitions of adversarial robustness or safety \cite{cui2024robustness}. For example, adversarial prompt injection can be used to subtly manipulate the model's behavior, while cross-modal adversarial attacks introduce perturbations across multiple modalities—either sequentially or jointly—to compromise the outputs integrity of the models \cite{wang2024stop}, \cite{yang2024can}.
\par
We summarize the existing techniques from the perspective of modalities in Fig. \ref{OverallTaxonomy}, categorizing them into attack and defense methods. Our analysis reveals significant imbalances in terms of the number of studies and the diversity of approaches for attacks and defenses. We hope that our work will inspire and contribute to future efforts aimed at enhancing the security of MLLMs.

\definecolor{11}{HTML}{ABC64E}
\definecolor{22}{HTML}{c39398}
\definecolor{33}{HTML}{FBB463}
\definecolor{44}{HTML}{80B1D3}
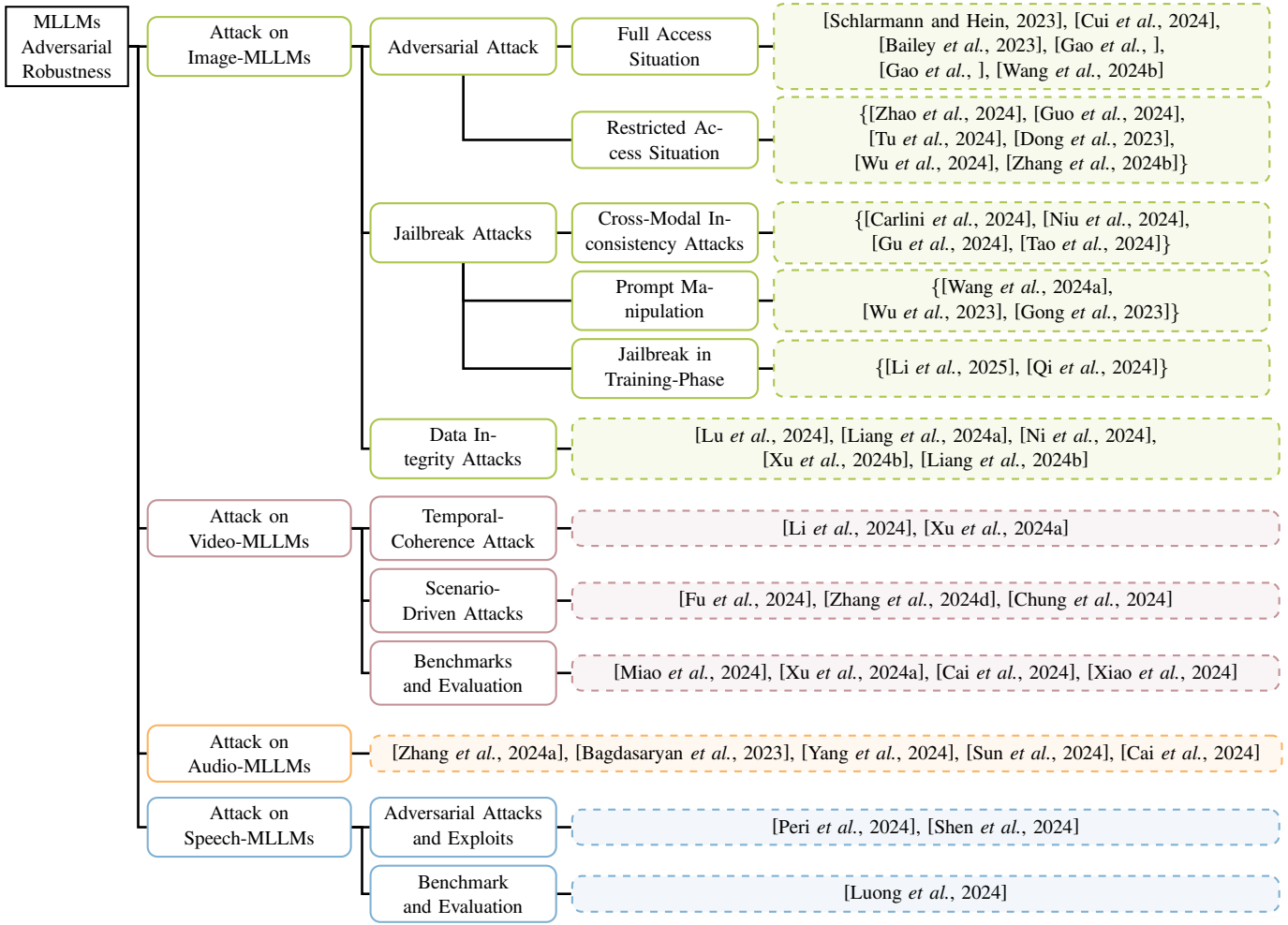
\begin{figure*}[!t]
\newcommand{\customfontsize}{\fontsize{8pt}{10pt}\selectfont}
\tikzset{
base/.style = {draw=red, thick, font=\customfontsize, rectangle},
root/.style = {base, minimum width=1.5 cm, minimum height=0.8 cm, fill=none, align=center, text width=1.5cm},
process-purple/.style = {base, minimum width=2.65 cm, minimum height=0.8cm, fill=none, rounded corners, align=center, text width=2.65 cm, draw=11},
process-red/.style = {base, minimum width=2.65 cm, minimum height=0.8cm, fill=none, rounded corners, align=center, text width=2.65 cm, draw=22},
process-orange/.style = {base, minimum width=2.65 cm, minimum height=0.8cm, fill=none, rounded corners, align=center, text width=2.65 cm, draw=33},
process-blue/.style = {base, minimum width=2.65 cm, minimum height=0.8cm, fill=none, rounded corners, align=center, text width=2.65 cm, draw=44},
process-purple-2/.style = {base, minimum width=2.4 cm, minimum height=0.8cm, fill=none, rounded corners, align=center, text width=2.4 cm, draw=11},	
process-purple-3/.style = {base, minimum width=2.4 cm, minimum height=0.84cm, fill=none, rounded corners, align=center, text width=2.4 cm, draw=11},
process-red-2/.style = {base, minimum width=2.4 cm, minimum height=0.56cm, fill=none, rounded corners, align=center, text width=2.4 cm, draw=22},
process-red-22/.style = {base, minimum width=2.4 cm, minimum height=0.9cm, fill=none, rounded corners, align=center, text width=2.4 cm, draw=22},
process-orange-2/.style = {base, minimum width=2.4 cm, minimum height=0.8cm, fill=none, rounded corners, align=center, text width=2.4 cm, draw=33},	
process-orange-3/.style = {base, minimum width=2.4 cm, minimum height=0.84cm, fill=none, rounded corners, align=center, text width=2.4 cm, draw=33},
process-blue-2/.style = {base, minimum width=2.4 cm, minimum height=0.8cm, fill=none, rounded corners, align=center, text width=2.4 cm, draw=44},	
process-blue-3/.style = {base, minimum width=2.4 cm, minimum height=0.84cm, fill=none, rounded corners, align=center, text width=2.4 cm, draw=44},
process-purple-m1/.style = {base, minimum width=1.5 cm, minimum height=0.56cm, fill=11!10, rounded corners, draw=22, dashed, align=center, text width=9.8 cm, draw=11},	
process-purple-m2/.style = {base, minimum width=1.5 cm, minimum height=0.8cm, fill=11!10, rounded corners, draw=22, dashed, align=center, text width=9.8 cm, draw=11},
process-red-m1/.style = {base, minimum width=1.5 cm, minimum height=0.5cm, fill=22!10, rounded corners, align=center, text width=9.8 cm, dashed, draw=22},
process-orange-m1/.style = {base, minimum width=1.5 cm, minimum height=0.5cm, fill=33!10, rounded corners, align=center, text width=9.8 cm, dashed, draw=33},
process-blue-m1/.style = {base, minimum width=1.5 cm, minimum height=0.5cm, fill=44!10, rounded corners, align=center, text width=9.8 cm, dashed, draw=44},
arrow/.style={black, line width=1pt}
}

\begin{tikzpicture}[node distance=0.25cm and 0.3cm, auto]	
    \node (MLLMs_AR) [root, minimum width=1.5cm, draw=black] {MLLMs\\Adversarial\\Robustness}; 
    \node (NodeAIM) [process-purple, right=0.25cm of MLLMs_AR] {Attack on Image-MLLMs};
        \node (NodeAIM_AA) [process-purple-2, right=0.25cm of NodeAIM] {Adversarial Attack};
            \node (NodeAIM_AA_Fully) [process-purple-2, right=0.2cm of NodeAIM_AA] {Full Access Situation};
                \node (NodeAIM_AA_Fully_M) [process-purple-m2, right=0.2cm of NodeAIM_AA_Fully, text width=6.8 cm] {\cite{schlarmann2023adversarial}, \cite{cui2024robustness},  \cite{baileyimage}, \cite{gaoinducing}, \cite{gaoinducing}, \cite{wang2024stop}};%
            \node (NodeAIM_AA_Partial) [process-purple-2, below=0.5cm of NodeAIM_AA_Fully] {Restricted Access Situation};
                \node (NodeAIM_AA_Partial_M) [process-purple-m2, right=0.2cm of NodeAIM_AA_Partial, text width=6.8 cm] {\{\cite{zhao2024evaluating}, \cite{guo2024efficient}, \cite{HowManyAre}, \cite{dong2023robust}, \cite{wu2024dissecting}, \cite{zhang2024b}\}};%
            
        \node (NodeAIM_JA) [process-purple-3, below=1.8cm of NodeAIM_AA] {Jailbreak Attacks};
            \node (NodeAIM_JA_CMIA) [process-purple-2, right=0.2cm of NodeAIM_JA] {Cross-Modal Inconsistency Attacks};
                \node (NodeAIM_JA_CMIA_M) [process-purple-m2, right=0.2cm of NodeAIM_JA_CMIA, text width=6.8 cm]{\{\cite{carlini2024aligned}, \cite{niu2024jailbreaking},
                \cite{guagent}, \cite{tao2024imgtrojan}\}};
                
            \node (NodeAIM_JA_PM) [process-purple-2, below=0.1cm of NodeAIM_JA_CMIA] {Prompt Manipulation};
                \node (NodeAIM_JA_PM_M) [process-purple-m2, right=0.2cm of NodeAIM_JA_PM, text width=6.8 cm]{\{\cite{wang2024white}, \cite{wu2023jailbreaking}, \cite{gong2023figstep}\}};
                
            \node (NodeAIM_JA_JTP) [process-purple-2, below=0.1cm of NodeAIM_JA_PM] {Jailbreak in Training-Phase};
                \node (NodeAIM_JA_JTP_M) [process-purple-m2, right=0.2cm of NodeAIM_JA_JTP, text width=6.8 cm]{\{\cite{li2025images}, \cite{qi2024visual}\}};
            
        \node (NodeAIM_DIA) [process-purple-3, below=2.2cm of NodeAIM_JA] {Data Integrity Attacks};
            \node (NodeAIM_DIA_M) [process-purple-m1, right=0.2cm of NodeAIM_DIA, minimum width=200]{\cite{lu2024test}, \cite{liang2024vltrojan}, \cite{ni2024physical}, \cite{xu2024shadow}, \cite{liang2024revisit}};
            
    \node (NodeAVM) [process-red, below=6.0cm of NodeAIM] {Attack on Video-MLLMs};
        \node (NodeAVM_AA_TCA) [process-red-22, right=0.25cm of NodeAVM] {Temporal-Coherence Attack};
            \node (NodeAVM_AA_TCA_M) [process-red-m1, right=0.2cm of NodeAVM_AA_TCA, minimum width=200]{\cite{li2024fmm}, \cite{xu2024beyond}};
        \node (NodeAVM_AA_SDA) [process-red-22, below=0.1cm of NodeAVM_AA_TCA] {Scenario-Driven Attacks};
            \node (NodeAVM_AA_SDA_M) [process-red-m1, right=0.2cm of NodeAVM_AA_SDA, minimum width=200]{\cite{fu2024pg}, \cite{zhang2024visual}, \cite{chung2024towards}};
        \node (NodeAVM_AA_BE) [process-red-22, below=0.1cm of NodeAVM_AA_SDA] {Benchmarks and Evaluation};
            \node (NodeAVM_AA_BE_M) [process-red-m1, right=0.2cm of NodeAVM_AA_BE,  minimum width=200]{\cite{miaot2vsafetybench}, \cite{xu2024beyond}, \cite{cai2024av}, \cite{xiao2024videoqa}};

    \node (NodeAAM) [process-orange, below=9.2cm of NodeAIM] {Attack on Audio-MLLMs};
        \node (NodeAAM_AA_M) [process-orange-m1, right=0.25cm of NodeAAM, text width=12.7 cm]{\cite{zhang2024badrobot}, \cite{bagdasaryan2023ab}, \cite{yang2024can}, \cite{sun2024trustnavgpt}, \cite{cai2024av}};
    \node (NodeASM) [process-blue, below=0.2cm of NodeAAM] {Attack on Speech-MLLMs};
        \node (NodeASM_AAE) [process-blue-2, right=0.25cm of NodeASM] {Adversarial Attacks and Exploits};
            \node (NodeASM_AAE_M) [process-blue-m1, right=0.2cm of NodeASM_AAE,  minimum width=200]{\cite{peri2024speechguard}, \cite{shen2024voice}};
        \node (NodeASM_BE) [process-blue-2, below=0.1cm of NodeASM_AAE] {Benchmark and Evaluation };
            \node (NodeASM_BE_M) [process-blue-m1, right=0.2cm of NodeASM_BE,  minimum width=200]{\cite{luong2024llamapartialspoof}};
    \coordinate (RightCoord_MLLMs_AR) at ([xshift=1.01cm]MLLMs_AR);
    \coordinate (BranchPoint_MLLMs_AR) at ([xshift=-0cm]MLLMs_AR.east);
    
    \coordinate (RightCoord_NodeAIM) at ([xshift=1.6cm]NodeAIM);
    \coordinate (BranchPoint_NodeAIM) at ([xshift=0cm]NodeAIM.east);
    
    \coordinate (RightCoord_NodeAVM) at ([xshift=1.6cm]NodeAVM);
    \coordinate (BranchPoint_NodeAVM) at ([xshift=0cm]NodeAVM.east);
    
    \coordinate (RightCoord_NodeAAM) at ([xshift=1.6cm]NodeAAM);
    \coordinate (BranchPoint_NodeAAM) at ([xshift=0cm]NodeAAM.east);

    \coordinate (RightCoord_NodeASM) at ([xshift=1.6cm]NodeASM); 
    \coordinate (BranchPoint_NodeASM) at ([xshift=0cm]NodeASM.east);	
    \draw [arrow] (MLLMs_AR.east) -- (BranchPoint_MLLMs_AR);		
    \draw [arrow] (BranchPoint_MLLMs_AR) -- (RightCoord_MLLMs_AR) |- (NodeAIM);
    \draw [arrow] (BranchPoint_NodeAIM) -- (RightCoord_NodeAIM) |- (NodeAIM_AA);
    \draw [arrow] (BranchPoint_NodeAIM) -- (RightCoord_NodeAIM) |- (NodeAIM_JA);
    \draw [arrow] (BranchPoint_NodeAIM) -- (RightCoord_NodeAIM) |- (NodeAIM_DIA);
    
    \draw [arrow] (NodeAIM_AA) -- (NodeAIM_AA_Fully) -- (NodeAIM_AA_Fully_M);	
    \draw [arrow] (NodeAIM_AA) |- (NodeAIM_AA_Partial) -- (NodeAIM_AA_Partial_M);
    
    \draw [arrow] (NodeAIM_JA) -- (NodeAIM_JA_CMIA) -- (NodeAIM_JA_CMIA_M);
    \draw [arrow] (NodeAIM_JA) |- (NodeAIM_JA_PM) -- (NodeAIM_JA_PM_M);
    \draw [arrow] (NodeAIM_JA) |- (NodeAIM_JA_JTP) -- (NodeAIM_JA_JTP_M);
    
    \draw [arrow] (NodeAIM_DIA) -- (NodeAIM_DIA_M);
    \draw [arrow] (BranchPoint_MLLMs_AR) -- (RightCoord_MLLMs_AR) |- (NodeAVM);
    \draw [arrow] (BranchPoint_NodeAVM) -- (RightCoord_NodeAVM) -- (NodeAVM_AA_TCA);
    \draw [arrow] (BranchPoint_NodeAVM) |- (RightCoord_NodeAVM) |- (NodeAVM_AA_SDA);
    \draw [arrow] (BranchPoint_NodeAVM) |- (RightCoord_NodeAVM) |- (NodeAVM_AA_BE);
    
    \draw [arrow] (NodeAVM_AA_TCA) -- (NodeAVM_AA_TCA_M);
    \draw [arrow] (NodeAVM_AA_SDA) -- (NodeAVM_AA_SDA_M);
    \draw [arrow] (NodeAVM_AA_BE) -- (NodeAVM_AA_BE_M);
    \draw [arrow] (BranchPoint_MLLMs_AR) -- (RightCoord_MLLMs_AR) |- (NodeAAM);
    \draw [arrow] (BranchPoint_NodeAAM) -- (RightCoord_NodeAAM) -- (NodeAAM_AA_M);
    \draw [arrow] (BranchPoint_MLLMs_AR) -- (RightCoord_MLLMs_AR) |- (NodeASM);
    \draw [arrow] (BranchPoint_NodeASM) -- (RightCoord_NodeASM) -- (NodeASM_AAE);
    \draw [arrow] (BranchPoint_NodeASM) -- (RightCoord_NodeASM) |- (NodeASM_BE);
    \draw [arrow] (NodeASM_AAE) -- (NodeASM_AAE_M);
    \draw [arrow] (NodeASM_BE) -- (NodeASM_BE_M);
\end{tikzpicture}
\caption{A Taxonomy of Adversarial Robustness of MLLMs.}
\label{OverallTaxonomy}
\end{figure*}

\section{Attack on Image-MLLMs}
This section provides a taxonomy and overview of the methods employed to exploit the vulnerabilities of MLLMs within the image modality, specifically focusing on adversarial attacks, jailbreak attacks, and data integrity attacks.
\subsection{Adversarial Attack}
Adversarial attacks pose robustness risks to MLLMs by manipulating inputs to induce incorrect outputs. Following, we categorize the scenarios based on the level of access to model information into two types: full access to model details and restricted access to partial information.
\subsubsection{Full Access Situation}
By exploiting the vulnerabilities of visual encoders, several works have extended traditional adversarial attacks to MLLMs. Schlarmann $et~al.$ conduct both untargeted and targeted attacks on MLLMs by introducing carefully crafted adversarial perturbations. Results demonstrate that subtle perturbations on visual inputs are shown to induce misclassifications or misleading outputs, which highlights the requirement for effective defensive strategies to address significant performance degradation caused by adversarial perturbations \cite{schlarmann2023adversarial}. Cui et al. assess adversarial robustness in multiple vision-related tasks, revealing that adversarial perturbations in the visual domain significantly degrade model performance on image classification and captioning tasks. However, for visual question answering, the models demonstrate greater adversarial robustness, with only slight performance degradation observed. Building on these findings, they propose contextual enhancements and task decomposition as promising strategies to mitigate these vulnerabilities \cite{cui2024robustness}. 
\par
Another concept, image hijacking, manipulates model behavior through subtle visual perturbations. This method excels in scenarios such as misinformation dissemination and jailbreak attacks, outperforming traditional adversarial techniques and exposing critical risks for content moderation applications \cite{baileyimage}. The Verbose Images is presented in \cite{gaoinducing}, which leverages imperceptible perturbations to induce excessively verbose outputs in MLLMs, significantly increasing energy consumption and latency during inference. This stealthy and transferable attack undermines computational efficiency, posing substantial risks to the availability of VLM-based systems \cite{gaoinducing}. 
\par
Adversarial vulnerabilities also extend to reasoning mechanisms within MLLMs. Wang et al. investigate the impact of Chain-of-Thought (CoT) reasoning on the adversarial robustness of MLLMs, revealing that while CoT improves robustness against existing attack methods through multi-step reasoning, the improvement is limited. Accordingly, the Stop-Reasoning Attack is proposed, targeting CoT reasoning frameworks by prematurely terminating reasoning processes or introducing errors. By bypassing CoT’s intermediate reasoning steps \cite{wang2024stop}.

\subsubsection{Restricted Access Situation}
While full access to model details allows for a comprehensive analysis of MLLM vulnerabilities to adversarial attacks, real-world commercial MLLMs often restrict access to model information. To examine vulnerabilities under more practical conditions, adversarial attacks in limited-information settings exploit this restricted visibility to execute manipulations. These methods typically operate under black- or gray-box assumptions, relying on external queries or surrogate models to generate adversarial inputs.
\par
To achieve black-box attacks, transfer-based approaches evaluate adversarial robustness under black-box settings. By utilizing surrogate models such as CLIP and BLIP, these methods enable the generation of adversarial examples that effectively transfer across various vision-language models, including MiniGPT-4 and BLIP-2 \cite{zhao2024evaluating}. Leveraging diffusion models to generate targeted and highly transferable adversarial examples for MLLMs provides a comprehensive assessment of potential security vulnerabilities before deployment. AdvDiffVLM enhances transfer-based attacks by employing ensemble gradient estimation and mask generation techniques to produce visually natural adversarial examples with dispersed semantics, thereby improving transferability \cite{guo2024efficient}. 
\par
In addition to conventional transfer-based attacks for achieving black-box settings, some studies have shifted focus toward evaluating the security of MLLMs within black-box scenarios. Adopting the Unicorn safety evaluation framework provides a systematic evaluation of the vulnerabilities of MLLMs under out-of-distribution and adversarial conditions. This benchmark tests MLLMs against attack scenarios and unconventional inputs, revealing robustness and safety protocol gaps. The findings suggest that visual language training pipelines often weaken original safety measures, increasing risks during practical use \cite{HowManyAre}. Studies on Google Bard reveal that targeted adversarial image and text attacks can bypass safety mechanisms, including face privacy detection and toxic content filters, leading to harmful outputs. The results present the critical need for more robust safety evaluations and defense strategies for multimodal systems \cite{dong2023robust}. The agent robustness evaluation framework models autonomous agents powered by MLLMs as interconnected components, systematically analyzing their vulnerabilities to adversarial inputs. Corresponding results identify critical components, such as evaluators and policy models, as particularly susceptible, with attacks achieving high success rates even with minimal perturbations \cite{wu2024dissecting}. Lastly, the B-AVIBench framework addresses the vulnerabilities of MLLMs to black-box adversarial visual instructions, spanning diverse attack methods across image, text, and biased content modalities. It systematically reveals significant security gaps, including susceptibility to multimodal biases and combined modality attacks, emphasizing the need for robust and fair defense mechanisms. Besides, this framework provides a comprehensive evaluation tool for open-source and proprietary models \cite{zhang2024b}.

\subsection{Jailbreak Attacks}
Jailbreak attacks bypass the safety mechanisms or constraints implemented in models by exploiting underlying system vulnerabilities. These attacks manipulate MLLMs to generate harmful or unauthorized outputs. In this section, we review the advancements in jailbreak attacks in MLLMs, with the first part focusing on general jailbreak attacks and the second examining prompt injection techniques.
\subsubsection{Cross-Modal Inconsistency Attacks}
Cross-modal attacks exploit alignment gaps between modalities. As a representative work, \cite{carlini2024aligned} craft adversarial prompts to manipulate attention mechanisms, bypassing safety constraints through multimodal perturbations. These adversarial inputs manipulate the model's behavior, enabling it to ignore safety constraints and generate harmful content. \cite{niu2024jailbreaking} further reveal that contradictory inputs across modalities (e.g., conflicting images and text) induce model misinterpretations, enabling jailbreaks transferable across MiniGPT-v2 and LLaVA. Similarly, by exploiting cross-modal inconsistencies, a variant termed ``infectious jailbreak'' allows compromised agents to propagate harmful behaviors across interconnected systems, threatening large-scale deployments \cite{guagent}. Additionally, \cite{tao2024imgtrojan} design adversarial examples that replace textual captions with malicious prompts, exploiting fusion layer vulnerabilities to generate harmful responses.

\subsubsection{Prompt Manipulation}
\cite{wang2024white} propose a white-box strategy using a ``Universal Master Key'', which combines adversarial image prefixes and text suffixes to bypass alignment defenses with high success rates. Conversely, \cite{wu2023jailbreaking} demonstrate that optimizing system prompts can enhance attack efficacy across languages, yet properly designed prompts also mitigate risks. Visual prompt injection further amplifies threats. \cite{gong2023figstep} embed harmful instructions into typographic images, breaking safety filters of GPT-4V.

\subsubsection{Jailbreak in Training-Phase}
\cite{li2025images} investigate how the presence of adversarial examples in training data can create vulnerabilities in MLLMs. The results argue that adversarial examples during training can expose gaps in the model’s ability to handle unexpected inputs, thereby facilitating successful jailbreak attempts. Similarly, Qi $et~al.$ delve deeper into visual prompt injection using adversarial examples, demonstrating subtle manipulations of visual inputs can bypass safety mechanisms in MLLMs \cite{qi2024visual}. Their findings reveal that visual prompts can effectively compel models to produce unsafe outputs, such as hate speech and violent instructions.

\subsection{Data Integrity Attacks}
Data integrity attacks deliberately manipulate data to compromise the training or operation of MLLMs. These attacks significantly impact the security and reliability of MLLMs, particularly in scenarios involving privacy-sensitive or high-stakes tasks. Data integrity attacks can be broadly classified into backdoor attacks and poisoning attacks.
\par
An early exploration of backdoor attacks for MLLMs is AnyDoor, which introduces a test-time backdoor attack against MLLMs, which decouples the backdoor setup by using universal adversarial perturbations applied to visual inputs, while activation is triggered by specific textual inputs. This separation enhances the attack's stealth and effectiveness. AnyDoor demonstrates remarkable adaptability, achieving a $98.5\%$ attack success rate across multiple MLLMs, even under challenging conditions such as noise interference and cropping \cite{lu2024test}. After that, VL-Trojan extends backdoor attack capabilities by integrating visual and textual triggers during instruction tuning. Leveraging feature isolation and clustering, this technique achieves over $99\%$ success rates with only poisoning $0.5\%$ of examples. Its consistent effectiveness across models and tasks demonstrates the elevated risk of backdoor vulnerabilities in instruction-tuned MLLMs \cite{liang2024vltrojan}. BadVLMDriver, the first backdoor attack targeting MLLMs in autonomous driving, explores physical backdoor attacks in real-world environments. By leveraging natural triggers, such as red balloons, BadVLMDriver can induce harmful behaviors, such as sudden acceleration, with a high attack success rate. This approach offers significant flexibility in selecting both the trigger and the resulting malicious behavior, thereby enhancing the stealth and practicality of the attack in diverse real-world scenarios \cite{ni2024physical}. Further studies reveal the cross-domain robustness of backdoor attacks, demonstrating that triggers embedded during instruction tuning can sustain success rates of over $97\%$, even when subjected to domain shifts. This proposes the requirement for effective detection and mitigation strategies, thereby emphasizing the escalating security risks posed by the increasing generalizability of backdoors in MLLMs \cite{xu2024shadow}.
\par
For data poisoning, Shadowcast exemplifies the sophistication of such attacks. It utilizes clean-label strategies to subtly alter image-text pairs, thereby manipulating model outputs without introducing obvious disruptions. Experimental results demonstrate Shadowcast's effectiveness, achieving attack success rates exceeding $80\%$ with fewer than $50$ poisoned examples while maintaining robustness across various data augmentations \cite{liang2024revisit}.

\section{Attack on Video-MLLMs}
Video-MLLMs integrate temporal and cross-modal reasoning for tasks like video QA and autonomous driving, yet their dynamic nature amplifies adversarial risks \cite{fu2024pg}. Adversarial strategies against video-MLLMs often target their temporal reasoning and cross-modal alignment \cite{li2024fmm}  \cite{xu2024beyond}. To address these challenges, recent benchmarks evaluate robustness across multiple safety dimensions \cite{miaot2vsafetybench}. Below, we provide a review of three aspects.

\subsection{Temporal-Coherence Attack}
\cite{li2024fmm} propose FMM-Attack, injecting optical flow-guided perturbations into keyframes to disrupt temporal dynamics. FMM-Attack degrades output accuracy even with sparse frame modifications, and joint attacks on video and language modalities are better than single-modality situations. Cross-modal threats further exploit alignment gaps: \cite{xu2024beyond} demonstrate that edited videos (e.g., social media deepfakes) mislead models by synchronizing deceptive visual and textual cues.

\subsection{Scenario-Driven Attacks}
MLLMs-based autonomous driving scenarios also face threats from adversarial attacks. \cite{fu2024pg} design PG-Attack, which combines precision-masked perturbations targeting critical regions such as vehicle movement, pedestrian activity, and traffic light changes with deceptive text patches to mislead model reasoning. Evaluated on CARLA-generated data, PG-Attack maintains perceptual stealth while achieving high attack success rates. Similarly, \cite{zhang2024visual} propose ADvLM, which optimizes perturbations on key frames using semantic-invariant prompts, ensuring robustness against viewpoint and lighting changes. To enable transferable attacks against Vision-MLLMs in autonomous driving, \cite{chung2024towards} embed misleading typographic text into video frames such as altered traffic signs.

\subsection{Benchmarks and Evaluation}
Additionally, existing efforts aim to evaluate the robustness of video-MLLMs from various perspectives. In text-to-video generative models, T2VSafetyBench \cite{miaot2vsafetybench} assesses generative models against explicit content and misinformation, revealing higher risks in advanced models due to enhanced semantic understanding. Benchmark studies EditVid-QA \cite{xu2024beyond} and AV-Deepfake1M \cite{cai2024av} benchmark performance on edited videos and synthetic deepfakes, respectively. Empirical studies \cite{xiao2024videoqa} further expose deficiencies in temporal grounding and adversarial robustness.

\section{Attack on Audio-MLLMs}
BadRobot exposes vulnerabilities in embodied MLLM that rely on audio inputs for interaction. Specifically, BadRobot analyzes three attacks, covering contextual jailbreaks, where carefully crafted audio inputs bypass restrictions to trigger harmful actions; safety misalignments, which exploit discrepancies between audio instructions and model responses; and conceptual deception, which breaks ethical constraints through subtasks embedded in speech instructions. Corresponding results reveal that even advanced systems like GPT-4-turbo exhibit weaknesses in these scenarios \cite{zhang2024badrobot}. The potential for indirect instruction injection through adversarial audio perturbations further underscores the sensitivity of speech modalities. These attacks leverage carefully crafted audio cues to influence the model’s output or alter its dialog flow. Experiments demonstrate over $90\%$ attack success rates in LLaVA \cite{bagdasaryan2023ab}. The Chat-Audio Attacks benchmark provides a structured framework for evaluating the robustness of MLLMs against diverse audio adversarial scenarios, including content modifications, emotional tone shifts, and implicit noise. Results reveal vulnerabilities that adversarial audio disrupts the semantic and contextual coherence of the models, even under controlled conditions \cite{yang2024can}. In response to these challenges, TrustNavGPT emerges to model audio uncertainties to improve the robustness of audio-guided navigation tasks. By integrating tonal ambiguities and pauses into its decision-making process, TrustNavGPT demonstrates a pathway toward mitigating adversarial risks in dynamic, real-world environments \cite{sun2024trustnavgpt}. Finally, the AV-Deepfake1M dataset emphasizes the importance of adversarial robustness in detecting temporally aligned manipulations across audio-visual modalities \cite{cai2024av}. Experiments on this dataset show significant performance drops in existing frameworks, pushing for advancements in training methodologies that consider the temporal and multimodal intricacies of speech-based inputs.

\section{Attack on Speech-MLLMs}
Speech data is dynamic and prone to perturbations, including adversarial noise, semantic manipulations, and spoofing attacks. As MLLMs are increasingly utilized in real-world applications, such as speech-driven question answering (QA) and audio-command systems, ensuring their robustness against adversarial threats is important.

\subsection{Adversarial Speech Attacks and Exploits}
Recent research indicates that speech-based MLLMs are highly vulnerable to advanced adversarial techniques. For instance, SpeechGuard \cite{peri2024speechguard} shows that Projected Gradient Descent can trigger unsafe responses, while its Time-Domain Noise Flooding defense markedly reduces attack success rates. For Voice Jailbreak Attacks, adversaries deploy narrative-based prompts to circumvent security measures in advanced models like GPT-4o \cite{shen2024voice}.

\subsection{Benchmark and Evaluation}
One benchmark evaluation in this area is the LlamaPartialSpoof dataset, which investigates model robustness against partial and fully spoofed speech \cite{luong2024llamapartialspoof}. Using advanced Text-to-Speech, it generates realistic audio examples that simulate semantic-altering and splicing attacks. The study exposes significant weaknesses in detecting partial spoofs and requires diverse datasets and evaluation frameworks to bolster defenses against threats. 

\section{Defense Methods}
\subsection{Visual Modality Defenses}
MLLM-Protector addresses adversarial image threats via a modular framework, combining a harm detector with a response detoxifier to filter unsafe outputs without degrading performance \cite{pi2024mllm}. After that, MM-SafetyBench establishes standardized safety evaluation protocols for vision-language models, leveraging typographic and diffusion-generated adversarial images to expose vulnerabilities \cite{liu2025mm}. For CoT-enhanced models, Stop-Reasoning proposes adversarial training to mitigate attacks targeting intermediate reasoning steps, improving alignment between visual and textual reasoning \cite{wang2024stop}. In parallel, FARE introduces unsupervised adversarial fine-tuning of vision encoders to preserve clean-task performance while hardening models against imperceptible perturbations, enabling robust deployment without downstream model adaptation \cite{schlarmannrobust}.

\subsection{Audio and Speech Modality Defenses}
SpeechGuard injects lightweight noise into audio signals to disrupt adversarial cues, improving robustness in spoken QA systems \cite{peri2024speechguard}. TrustNavGPT enhances robustness in audio-guided navigation by modeling semantic and tonal uncertainties, achieving a credible navigation performance when suffering adversarial attacks \cite{sun2024trustnavgpt}.

\subsection{Video Modality Defenses}
To enhance the robustness of video-LLMs, \cite{xiao2024videoqa} improves the strength of temporal reasoning, visual grounding, and adversarial resilience, ensuring that models effectively leverage visual information rather than exploiting linguistic shortcuts. Meanwhile, EditVid-QA \cite{xu2024beyond}, strengthens defenses against edited and manipulated videos by introducing a diverse benchmark covering effects, memes, and social media transformations. These defenses are realized through frame-consistency learning, adaptive visual grounding, and domain-adaptive fine-tuning, ensuring models learn from real visual data instead of exploiting linguistic shortcuts.

\section{Trends and Future Directions}
\subsection{Comprehensive Robustness Evaluation} 
Current robustness evaluation benchmarks are limited and can not be used to fully spectrum adversarial risks across modalities. Future efforts should prioritize the development of a multidimensional safety taxonomy that systematically defines safety dimensions. Additionally, there is a need to collect diverse and scalable datasets that balance volume, diversity, and adversarial realism. This includes incorporating synthetic adversarial examples (e.g., AdvDiffVLM-generated perturbations) and real-world attack scenarios (e.g., BadRobot’s audio jailbreaks). To further enhance evaluation metrics, advanced LLMs should be leveraged to automate safety assessments, addressing prompt sensitivity and hallucination risks.
\subsection{Cross-Modal and Transferable Attacks} While existing attacks expose modality-specific vulnerabilities, several gaps remain. Current attacks typically perturb individual modalities, but the interaction between perturbations (e.g., adversarial videos with synchronized audio-text triggers) remains underexplored. Moreover, most attacks assume white-box access or task-specific setups, limiting their applicability. Although methods such as AdvDiffVLM’s ensemble gradient estimation and CroPA’s cross-prompt adversarial transferability show promise, broader generalization to commercial models is necessary. Furthermore, video-MLLMs are vulnerable to temporal coherence exploits.
\subsection{Data-Centric Security Risk} The reliance of MLLMs on multimodal training data introduces unique risks. Despite advancements in clean-label poisoning and universal triggers, defenses against domain-shifted triggers remain insufficient. Additionally, adversarial attacks often exploit inherent biases in training data. Future research should quantify the interaction between biases and adversarial manipulations, developing debiasing techniques during model alignment. Optimizing visual instruction tuning with safety-aware datasets and reinforcement learning could reduce vulnerabilities. However, challenges such as reward misalignment and multimodal preference modeling require further exploration.

\subsection{Human-AI Collaborative Defense} As attacks on MLLMs become increasingly sophisticated, emerging defense paradigms integrate human expertise with automated systems. For example, TrustNavGPT \cite{sun2024trustnavgpt} leverages uncertainty in audio-guided navigation, but broader frameworks could benefit from human oversight to identify subtle adversarial patterns. Developing interpretable defenses is essential to enhance transparency and trust in safety mechanisms.

\subsection{Lacking Benchmarking and Standardization} 
Another obstacle to progress is the lack of standardized evaluation protocols. Existing attacks are tested on disparate models with inconsistent metrics. To address this, community-driven efforts should establish baselines for evaluation settings. Notably, current defenses mainly focus on modality-specific robustness and lack holistic metrics for cross-modal safety. Releasing modular frameworks will streamline research on attack or defense paradigms and foster further advancements in the field.

\section{Conclusion}
This survey explores the adversarial robustness of multimodal large language models (MLLMs). We begin by introducing the background and evaluation methods used to assess the robustness of MLLMs. We then categorize adversarial attacks across image, video, audio, and speech modalities, reviewing the vulnerabilities of MLLMs. Finally, we highlight key research gaps and unresolved issues that warrant further investigation in the field of MLLM robustness. We hope this survey provides valuable insights to researchers and encourages greater contributions to this growing area of study.

\bibliographystyle{named}
\bibliography{ijcai25}

\end{document}